\documentclass[letterpaper]{article}
\usepackage[preprint]{aaai2027}
\usepackage[hyphens]{url}
\usepackage{graphicx}
\usepackage{natbib}
\usepackage{caption}
\usepackage{booktabs}
\usepackage{multirow}
\usepackage{amsmath}
\usepackage{amssymb}

\newcommand{\cmark}{\ensuremath{\checkmark}}
\newcommand{\xmark}{\ensuremath{\times}}
\newcommand{\pmark}{part.}

\title{ReMMD: Realistic Multilingual Multi-Image Agentic Verification for Multimodal Misinformation Detection}
\author{
Chenhao Dang\textsuperscript{\rm 1,2},
Dantong Zhu\textsuperscript{\rm 4},
Jun Yang\textsuperscript{\rm 5},
Conghui He\textsuperscript{\rm 2},
Weijia Li\textsuperscript{\rm 2,3}\corresponding
}
\affiliations{
\textsuperscript{\rm 1}Shanghai Jiaotong University\quad
\textsuperscript{\rm 2}Shanghai Artificial Intelligence Laboratory\\
\textsuperscript{\rm 3}Tsinghua University\quad
\textsuperscript{\rm 4}Central South University\\
\textsuperscript{\rm 5}China Electronics Technology Group Corporation 15th Research Institute\\
dangchenhao@pjlab.org.cn, zhudantong@csu.edu.cn, yangjun15s@cetc.com.cn\\
heconghui@pjlab.org.cn, liweijia@sz.tsinghua.edu.cn
}

\begin{document}
\maketitle
\begin{abstract}
Multimodal misinformation detection is increasingly important because viral posts now combine long multilingual narratives, several images, mixed provenance, and subtle cross-modal framing errors. Existing benchmarks and methods remain poorly matched to this setting: they usually isolate short captions, single images, binary labels, or one manipulation source, while agentic verification remains costly under realistic evidence search. We present ReMMD, a realistic multilingual multi-image agentic verification framework for multimodal misinformation detection. ReMMD includes ReMMDBench, a real-world multimodal misinformation detection benchmark with 500 samples, 2,756 images, five monolingual evaluations, two cross-lingual settings, three text-length tiers, multi-image posts, five-way veracity labels, eight distortion labels, evidence provenance, and rationales. It also includes ReMMD-Agent, a persistent-memory verifier that decomposes posts into atomic points, builds a reusable evidence set, and predicts structured veracity verdicts, fine-grained distortion diagnoses, and explanatory rationales. Across proprietary systems, open LVLMs, MMD-Agent, and T$^2$-Agent, ReMMD-Agent obtains the best five-way veracity performance, with 41.80\% accuracy and 39.12\% macro-F1 using GPT-5.2, while reducing cost by 17.5\% relative to MMD-Agent and 79.9\% relative to T$^2$-Agent. The project is available at \url{https://dang-ai.github.io/ReMMD}.
\end{abstract}

\begin{figure}[t]
  \centering
  \includegraphics[width=0.98\columnwidth]{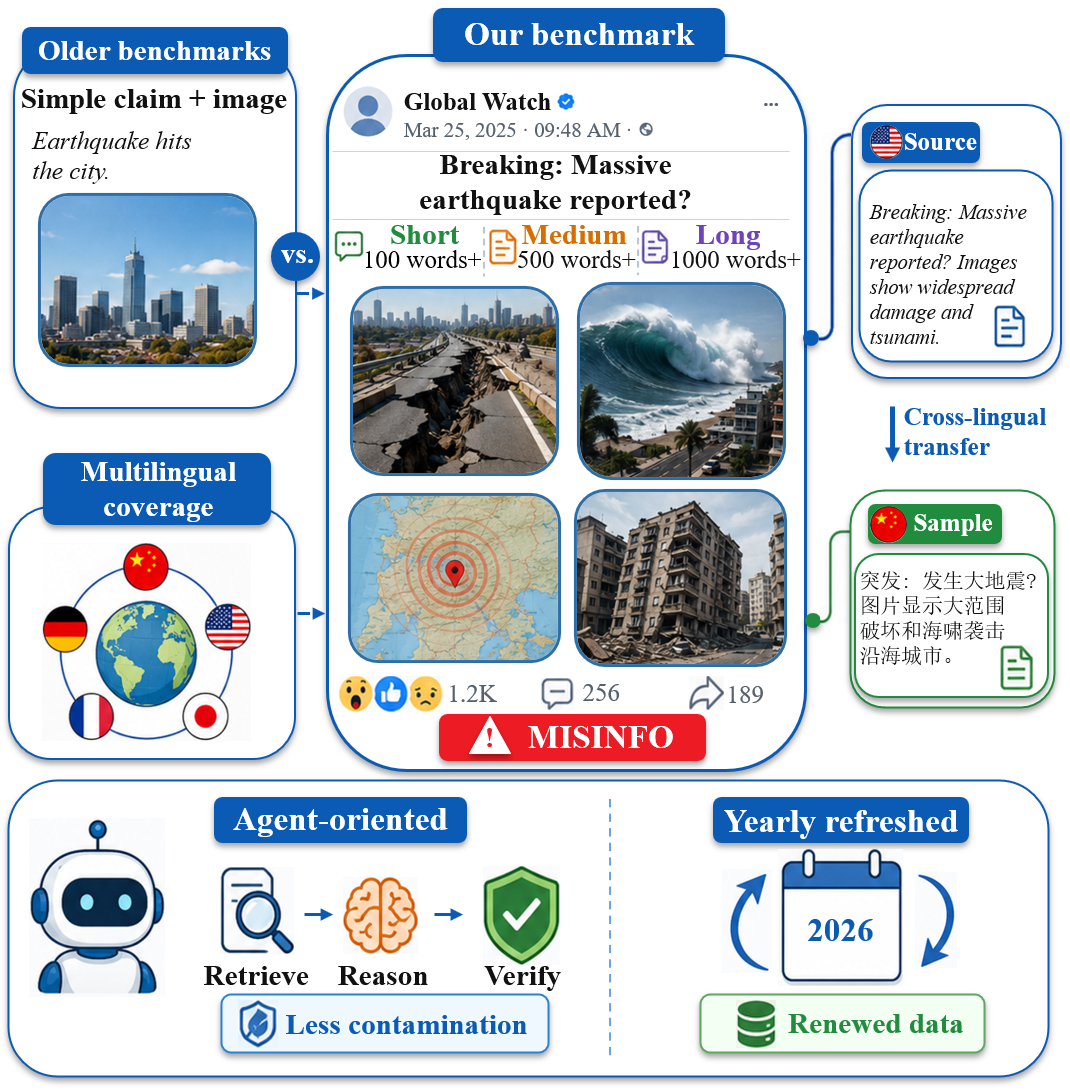}
  \caption{ReMMDBench is an agent-oriented benchmark for realistic multimodal misinformation detection, with yearly refreshed data to reduce contamination.}
  \label{fig:teaser}
\end{figure}

\begin{table*}[t]
\centering
\small
\setlength{\tabcolsep}{2.6pt}
\begin{tabular}{l|ccccccccc}
\toprule
\textbf{Benchmark} & \textbf{Dyn.} & \textbf{Agent} & \textbf{Lang.} & \textbf{Cross} & \textbf{Length} & \textbf{Images} & \textbf{Labels} & \textbf{Ratl.} & \textbf{Auto} \\
\midrule
NewsCLIPpings~\citep{luo2021newsclippings} & \xmark & \xmark & \xmark & \xmark & short & \xmark & 2 & \pmark & \xmark \\
MuMiN~\citep{nielsen2022mumin} & \xmark & \xmark & \cmark & \pmark & short & \xmark & 2 & \xmark & \xmark \\
COSMOS~\citep{aneja2021cosmos} & \xmark & \xmark & \xmark & \xmark & short & \xmark & 2 & \xmark & \xmark \\
VERITE~\citep{papadopoulos2023verite} & \xmark & \xmark & \xmark & \xmark & short & \xmark & 3 & \pmark & \pmark \\
AVeriTeC~\citep{cao2023averitec} & \pmark & \xmark & \xmark & \xmark & med. & \xmark & 4 & \cmark & \pmark \\
MFC-Bench~\citep{wang2025mfcbench} & \xmark & \xmark & \xmark & \xmark & med. & \pmark & 3 & \cmark & \pmark \\
GroundMM~\citep{yang2025grounding} & \xmark & \xmark & \xmark & \xmark & med. & \pmark & 2+g & \cmark & \pmark \\
MMFakeBench~\citep{liu2025mmfakebench} & \xmark & \cmark & \xmark & \xmark & short & \xmark & 2+3 & \cmark & \pmark \\
XFacta~\citep{xiao2025xfacta} & \pmark & \pmark & \xmark & \xmark & med. & \pmark & 2 & \cmark & \pmark \\
VeriTaS~\citep{rothermel2026veritas} & \cmark & \pmark & \pmark & \pmark & med. & \pmark & 4 & \cmark & \cmark \\
M4FC~\citep{geng2026m4fc} & \pmark & \xmark & \cmark & \cmark & med. & \pmark & multi & \cmark & \pmark \\
ReMMDBench (Ours) & \cmark & \cmark & \cmark & \cmark & all & \cmark & 5+8 & \cmark & \cmark \\
\bottomrule
\end{tabular}
\caption{Comparison with representative fact-checking and multimodal misinformation benchmarks. \textbf{Dyn.} indicates dynamically refreshed or update-aware data; \textbf{Agent} indicates whether designed for agentic verification; \textbf{Lang.} indicates multilingual coverage; \textbf{Cross} indicates cross-lingual evaluation; \textbf{Length} summarizes typical text length (short, med., long, or all tiers); \textbf{Images} indicates multi-image or image-aware samples; \textbf{Labels} gives the number or type of veracity and distortion labels, where ``2+g'' includes grounding supervision and ``5+8'' denotes five-way veracity plus eight distortion labels; \textbf{Ratl.} indicates rationale or explanation supervision; \textbf{Auto} indicates automated or dynamically assisted construction/evaluation. \cmark, \xmark, and \pmark denote yes, no, and partial support.}
\label{tab:benchmark-position}
\end{table*}

\begin{figure*}[t]
  \centering
  \includegraphics[width=0.98\textwidth]{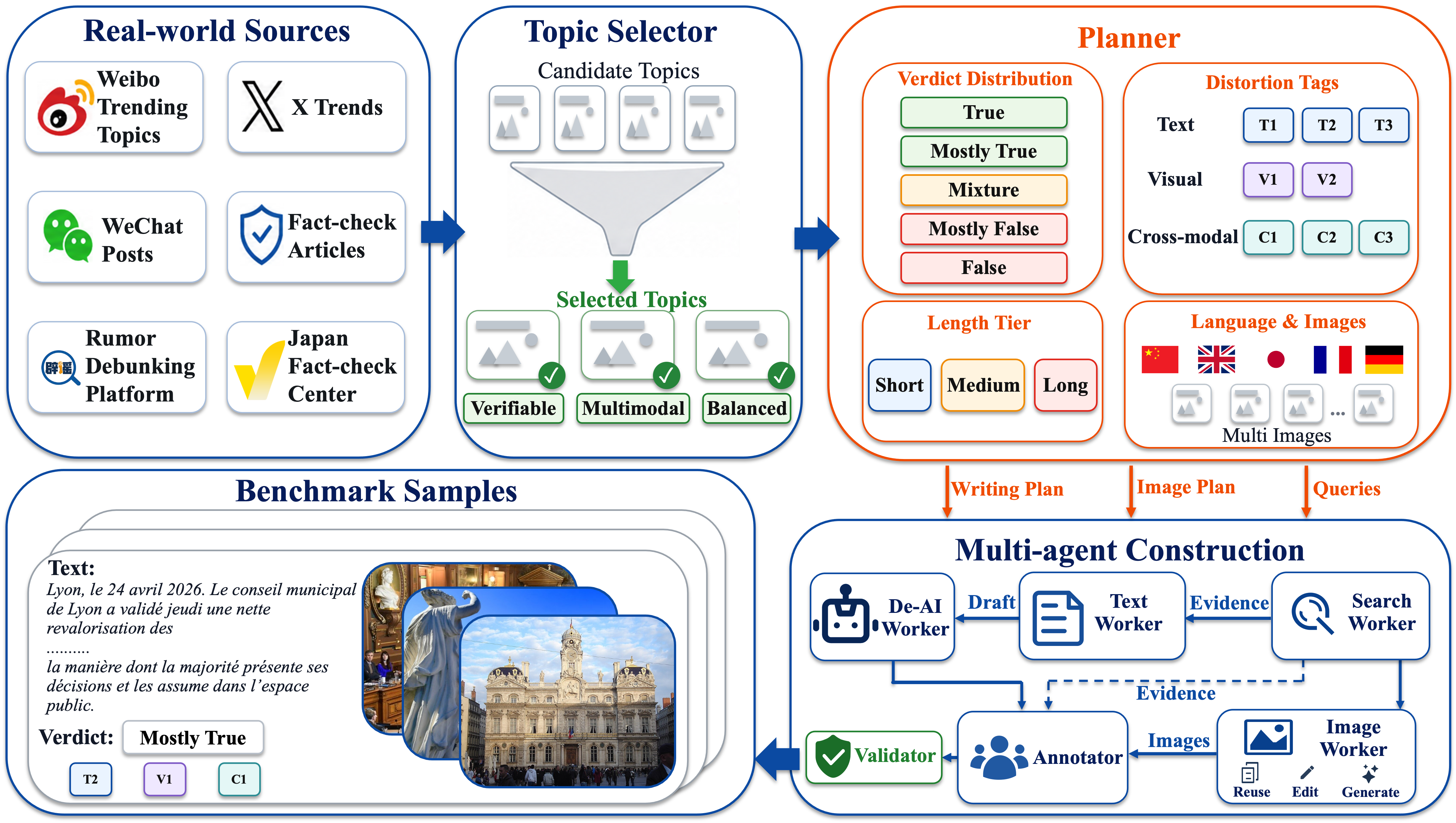}
  \caption{ReMMDBench turns real-world misinformation topics into controlled multilingual multi-image samples by planning language, length, visual provenance, and label conditions, then validating each instance against evidence, distortion annotations, text--image consistency, and image provenance before inclusion.}
  \label{fig:bench-pipeline}
\end{figure*}

\section{Introduction}

Real-world multimodal rumors and misinformation pervade news and social media, where text, images, screenshots, and generated or edited media jointly amplify false claims, threatening social trust, political processes, public figures, crisis response, and national security~\citep{vosoughi2018spread,lv2025multi}. This risk has shaped a progression of multimodal evaluation resources, from early cross-modal mismatch evaluation in NewsCLIPpings~\citep{luo2021newsclippings} to mixed-source and dynamically refreshed settings in MMFakeBench~\citep{liu2025mmfakebench} and VeriTaS~\citep{rothermel2026veritas}. Methods follow suit, with VLM and LVLM-based multimodal misinformation detection systems improving visual perception and retrieval~\citep{wang2025mfcbench,liu2025mmfakebench}, while T$^2$-Agent~\citep{cui2025t2agent} extends tool-augmented verification through search-based reasoning.

Nevertheless, benchmark-driven progress still leaves a gap between existing MMD evaluations and operational deployment. Existing evaluations often simplify verification to isolated claims, single images, coarse verdicts, or one manipulation source, while real deployments must handle long multilingual posts with many images, mixed visual provenance, partial truth, evolving evidence, and textual, visual, and cross-modal distortion attribution~\citep{giachanou2020multiimage,mullerbudack2020crossmodal}. Addressing this gap requires systems that can decompose central claims, select evidential images, track provenance, reuse evidence, and attribute distortions across modalities, and recent tool-augmented MMD and production generalist agents make such agentic verification increasingly feasible~\citep{cui2025t2agent,shlomov2026business}.

This gap motivates the Realistic Multimodal Misinformation Detection Benchmark (ReMMDBench), a real-world, agent-oriented benchmark for evaluating systems under operational verification conditions, as illustrated in Figure~\ref{fig:teaser} and positioned against prior resources in Table~\ref{tab:benchmark-position}. ReMMDBench consists of single-text, multi-image samples spanning three length tiers, five monolingual evaluations, and two cross-lingual transfer settings, as summarized in Table~\ref{tab:bench-stats}. Since real-world fact-checking often requires graded verdicts rather than binary truth labels~\citep{wang2017liar,lee2023factchecking}, each sample is annotated with a five-class L1 veracity label, L2 distortion labels selected from eight categories, and an L3 natural-language rationale.

Bridging this deployment gap under the same operational setting also calls for an agent that manages evidence before judgment. We introduce ReMMD-Agent, a real-world MMD verifier that decomposes posts into atomic claims and image bindings, retrieves web, image, and social evidence, and incrementally updates a persistent memory bank with reusable evidence. A structured judge then predicts L1 veracity, L2 distortion labels, and an L3 rationale from this evidence state. Like an experienced fact-checker, this workflow supports multidimensional judgment while remaining cost-efficient.

Together, ReMMDBench and ReMMD-Agent form ReMMD, a realistic multilingual multi-image agentic verification framework for MMD. On ReMMDBench, we evaluate two general-purpose closed-source agents and three open-source MMD agents, including ReMMD-Agent, using base models drawn from three backbone families and five total model sizes. ReMMD-Agent with GPT-5.2 sets the current best ReMMDBench result and reduces GPT-5.2 cost by 17.5\% relative to MMD-Agent and 79.9\% relative to T$^2$-Agent. Qwen3.5-9B also outperforms the closed-source agents on ReMMDBench and remains competitive on MMFakeBench. Our main contributions are summarized below.

\begin{itemize}
\item \textbf{ReMMDBench.} A real-world, agent-oriented MMD benchmark with multilingual, multi-image, varied-length samples, five-way veracity, eight distortion labels, and rationales.

\item \textbf{ReMMD-Agent.} A real-world MMD verifier that organizes claims, image bindings, and persistent evidence memory for low-cost structured judgment.

\item \textbf{Evaluation.} Broad comparisons across commercial agents and open agentic baselines validate ReMMDBench and show ReMMD-Agent's high performance, analyzing language, length, distortion, transfer, and cost.
\end{itemize}

\section{Related Work}

\paragraph{Datasets and benchmarks.} Multimodal misinformation benchmarks systematically evaluate visual manipulations and cross-modal discrepancies to identify image repurposing, unimodal bias, spatial localization, and semantic attribution~\citep{sabir2018repurposing,aneja2021cosmos,luo2021newsclippings,papadopoulos2023verite,yao2023endtoend,shao2023detecting,da2021edited,guo2024eachfake}. Recent benchmarks further cover AI-generated or edited news, mixed-source distortion, grounding, realism, multilinguality, and dynamic evaluation~\citep{huang2024miragenews,xu2024m3a,shu2024llmgen,li2025drifting,xu2025mdam3,liu2025mmfakebench,yang2025grounding,zhu2025mfnd,xiao2025xfacta,geng2026m4fc,rothermel2026veritas,nielsen2022mumin,li2020mmcovid}. Yet a massive gap remains between these evaluation settings and real-world deployment. Existing data typically isolates short captions, single images, limited languages, coarse labels, static evidence, or singular manipulation sources, whereas operational verification strictly demands resolving long multilingual posts, high-density images, mixed provenance, graded veracity, and fine-grained visual distortions under continuously evolving evidence conditions.

\paragraph{LVLMs and agentic verification.} Large vision-language models are natural multimodal verifiers, but recent studies show that perception alone remains vulnerable to grounding errors, stale or adversarial evidence, and temporal contamination~\citep{wang2025mfcbench,yang2025realfactbench,chen2026realtime,xu2026livefact,xuyan2026triplefact}. Agentic verification improves robustness by decomposing claims, asking targeted questions, and invoking retrieval or visual tools~\citep{beigi2025questions,liu2025mmfakebench,cui2025t2agent}; T$^2$-Agent further expands tool-augmented reasoning with Monte Carlo Tree Search, but the added search increases cost. Thus, the core method challenge is not only perception or tool access, but evidence management at deployment scale. Realistic MMD requires long-horizon memory over many claims, images, sources, timestamps, provenance cues, and contradictions to support accurate classification, while high-concurrency applications also demand strict control of repeated retrieval and inference cost.

\begin{table}[tbh]
\centering
\small
\setlength{\tabcolsep}{4pt}
\begin{tabular}{lr}
\toprule
\textbf{Statistic} & \textbf{Count} \\
\midrule
Samples & 500 \\
Images & 2,756 \\
Avg. images & 5.51 \\
Monolingual & 423 \\
Cross-lingual & 77 \\
AI-generated image & 237 \\
AI-edited image & 246 \\
Any AI-touched image & 384 \\
\midrule
Short / medium / long & 173 / 159 / 168 \\
English / Chinese & 111 / 112 \\
German / Japanese / French & 67 / 68 / 65 \\
\bottomrule
\end{tabular}
\caption{Core statistics of ReMMDBench. Length tiers are balanced, and most samples contain multiple images and at least one AI-touched visual item.}
\label{tab:bench-stats}
\end{table}

\section{ReMMDBench}

\subsection{Benchmark Design and Construction}

ReMMDBench is designed around controlled realism to make the verification pressures of multimodal misinformation strictly observable. As illustrated in Figure~\ref{fig:bench-pipeline}, the data collection methodology originates from a diverse seed set of real-world trending topics and verified debunking articles continuously scraped from platforms including Weibo, X, WeChat, and international fact-check centers. Let $S_{\mathrm{seed}}$ denote this collection of authentic raw posts. To construct isolated evaluation conditions without compromising real-world complexity, an automated multi-agent pipeline systematically transforms every authentic seed $s \in S_{\mathrm{seed}}$ into a controlled benchmark sample. A designated Planner agent dictates the target veracity, distortion configuration, language, and visual budget. Subsequently, specialized Text, Search, and Image Workers collaboratively draft the narrative, retrieve supporting evidence, and manipulate visual assets to engineer the prescribed structural misalignments.

\begin{table}[tbp]
\centering
\small
\setlength{\tabcolsep}{4pt}
\begin{tabular}{lrr}
\toprule
\textbf{Topic category} & \textbf{Count} & \textbf{Percent} \\
\midrule
Entertainment and sports & 107 & 21.4 \\
International conflict & 85 & 17.0 \\
Public safety and disaster & 70 & 14.0 \\
Science, technology, and AI & 66 & 13.2 \\
Politics and public affairs & 58 & 11.6 \\
Society and culture & 49 & 9.8 \\
Finance and markets & 35 & 7.0 \\
Health and medicine & 22 & 4.4 \\
Other & 8 & 1.6 \\
\bottomrule
\end{tabular}
\caption{Topic distribution of ReMMDBench. The benchmark avoids concentrating on a single rumor domain, which helps distinguish general verification ability from topic memorization.}
\label{tab:topics}
\end{table}

Table~\ref{tab:bench-stats} reports the main statistics. The benchmark is deliberately image-dense, with the average image count rising from 2.35 in short tiers to 10.05 in long tiers. To rigorously evaluate international rumor propagation, the benchmark introduces two distinct cross-lingual settings where native source material and corresponding visual contents are structurally migrated into a different target language. This pipeline systematically engineers semantic misalignments and contextual displacement during the translation process, creating sophisticated multimodal misinformation rooted in cross-cultural information shifts rather than straightforward fabrication. Short posts therefore test whether a model avoids over-reading compact claims, whereas long posts require tracking entities, dates, quotations, and image order across a larger visual context.

\subsection{Annotation Schema}

Each sample receives a hierarchical annotation comprising an L1 veracity label, L2 distortion labels, and an L3 natural-language rationale. The L1 labels are ordered by severity, distinguishing fully supported claims, minor local errors, mixed true and false evidence, dominantly false conclusions with residual true details, and unsupported core propositions. The middle labels encode whether an error changes the main conclusion or merely qualifies it. 

Informed by media-based typologies and mixed-source multimodal misinformation benchmarks~\citep{dufour2024ammeba,liu2025mmfakebench}, the L2 taxonomy rigorously separates textual, visual, and cross-modal distortions. Textual labels cover fabrication, distortion of a real factual basis, and misleading context. Visual labels cover synthetic content and editing. Cross-modal labels cover semantic, contextual, and pragmatic inconsistency. We deliberately separate visual provenance from evidential force, since authentic images remain misleading when attached to fabricated events.

\begin{figure*}[t]
  \centering
  \includegraphics[width=\textwidth]{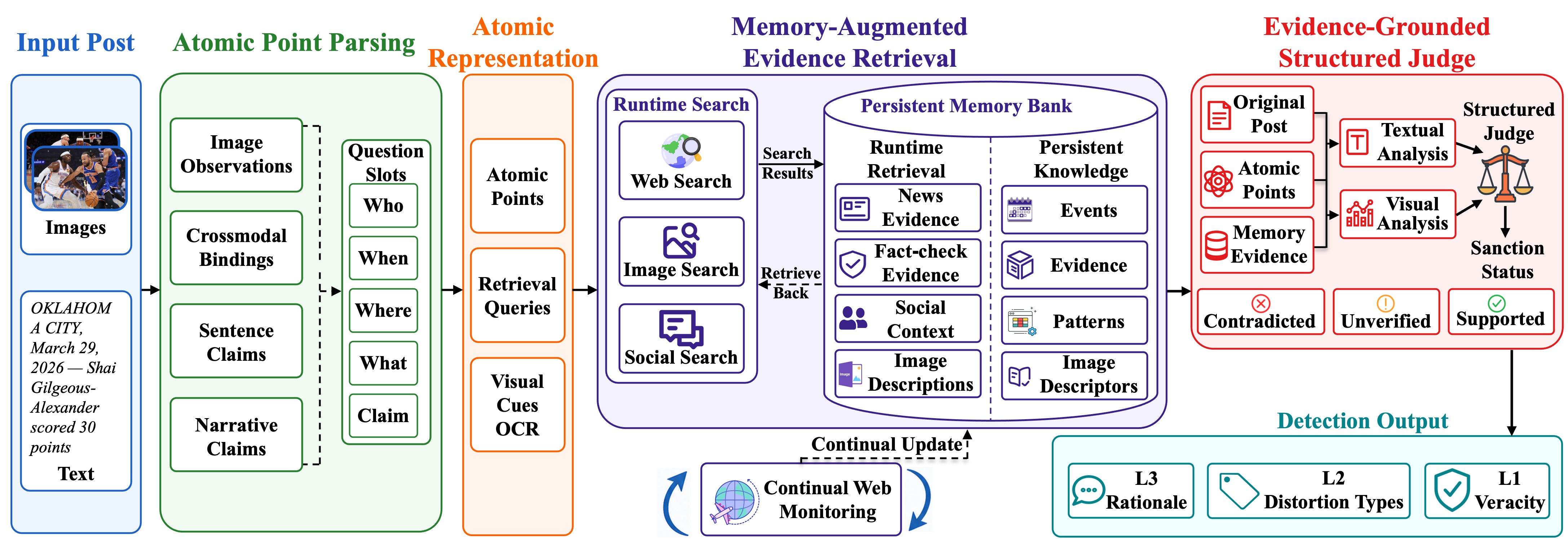}
  \caption{ReMMD-Agent verifies a multimodal post by first decomposing text and images into atomic claims, observations, and cross-modal bindings, then retrieving and reusing evidence in a persistent memory bank before a structured judge integrates textual, visual, and provenance cues to produce the L1 veracity label, L2 distortion diagnosis, and L3 rationale.}
  \label{fig:agent}
\end{figure*}

\subsection{Quality Control}

To ensure scalable and rigorous validation, ReMMDBench employs a fully automated quality control protocol supervised by a designated Validator agent, eliminating bottlenecks associated with massive human annotation while maintaining objective standards. The Validator systematically audits whether the generated L1 verdict logically follows from the synthesized central claim and verifies the physical grounding of all L2 distortion tags. To validate the reliability of this automated pipeline, we measured agreement against two independent professional human annotators over all 500 benchmark samples. This calibration yielded a strong Cohen's $\kappa$ of 0.91. Comprehensive human-AI consistency metrics including exact L1 accuracy distributions and detailed L2 exact match statistics are documented in Appendix~\ref{app:annotation-agreement} alongside exhaustive data cards. Furthermore, Table~\ref{tab:topics} reports the comprehensive topic mix. The benchmark strictly avoids concentration within a single rumor domain. Diverse categories including entertainment, conflict, public safety, science, politics, health, and finance demand entirely different evidence sources, actively testing both perceptual grounding and domain-sensitive retrieval. This deliberate breadth effectively limits shortcut learning, ensuring the same L1 verdict systematically arises from varied combinations of textual distortion, visual reuse, and cross-modal mismatch.

\section{ReMMD-Agent}

Figure~\ref{fig:agent} details the computation graph of ReMMD-Agent. Given an input post $s=(x,I)$ comprising textual content $x$ and a visual set $I=\{i_m\}_{m=1}^{M}$, the agent predicts a verdict tuple $(y,z,r)$ containing a five-way veracity label $y$, an eight-dimensional distortion vector $z\in\{0,1\}^8$, and a concise rationale $r$. Rather than judging the multimodal bundle directly, ReMMD-Agent compresses the input into verifiable atomic units, retrieves evidence around those units, and performs judgment over an explicit evidence state.

\subsection{Atomic Representation}

The initial stage applies a parsing transformation mapping the post $s$ into a set of atomic points
$$A=\{a_j=(c_j,q_j,v_j,\tau_j)\}_{j=1}^{n}$$
where $c_j$ constitutes a minimal claim or visual observation, $q_j$ formulates a retrieval query, $v_j$ isolates visual cues, and $\tau_j$ designates the point type. To standardize information extraction and align with the internal Question Slots designated in Figure~\ref{fig:agent}, the language model is constrained to populate predefined structural fields encompassing entity identities, temporal anchors, spatial locations, event descriptions, and core claims. The parser populates only the specific slots explicitly present in the local context while strictly disregarding absent variables. Near-duplicates are merged, restricting the retention to a maximum of twelve points per sample to reduce redundant searches while preserving critical classification evidence.

\subsection{Memory-Augmented Retrieval}

The retrieval stage anchors these points against a dynamic, sample-level memory bank $M_s=\{e_k\}_{k=1}^{K}$. Crucially, $M_s$ is asynchronously initialized and enriched by a Continual Web Monitoring module executing scheduled background sweeps across news aggregates and fact-checking domains to ingest emerging verified events. During active inference, the system utilizes the parsed queries $q_j$ and visual cues $v_j$ to execute targeted web and social tool calls. The agent evaluates the relevance of all integrated evidence $e \in M_s$ to the current atomic point $a_j$ utilizing cosine similarity over dense representations generated by a Qwen3-Embedding-8B encoder, yielding
$$R_j=\operatorname{TopK}_{e\in M_s}\operatorname{sim}(\phi(a_j),\phi(e))$$
where $\phi(\cdot)$ denotes the embedding function and the retrieval threshold $\operatorname{TopK}$ is rigorously set to 20. This persistent architecture functions as an auditable evidence state rather than a transient prompt context, seamlessly facilitating cross-claim evidence reuse.

\begin{table*}[t]
\centering
\small
\setlength{\tabcolsep}{2.2pt}
\begin{tabular}{llrrrrrrrr}
\toprule
\multirow{2}{*}{\textbf{Agent}} & \multirow{2}{*}{\textbf{Backbone}} & \multicolumn{4}{c}{\textbf{L1 Veracity}} & \multicolumn{4}{c}{\textbf{L2 Distortion}} \\
\cmidrule(lr){3-6}\cmidrule(lr){7-10}
 & & \textbf{Accuracy} & \textbf{Precision} & \textbf{Recall} & \textbf{Macro-F1} & \textbf{Precision} & \textbf{Recall} & \textbf{Macro-F1} & \textbf{Exact Match} \\
\midrule
Manus & proprietary & 33.00 & 33.58 & 33.00 & 33.13 & 43.69 & 42.32 & 42.75 & 7.60 \\
ChatGPT & proprietary & 30.20 & 32.65 & 30.15 & 28.24 & 41.71 & 47.56 & 43.63 & 3.00 \\
\midrule
\multirow{5}{*}{MMD-Agent} & GPT-5.2 & 26.40 & 23.77 & 26.38 & 23.42 & 42.24 & 48.83 & 41.98 & 2.00 \\
 & Gemma4-31B & 25.60 & 26.83 & 25.60 & 25.86 & 40.94 & 42.11 & 40.82 & 5.20 \\
 & Qwen3.6-27B & 25.40 & 26.30 & 25.44 & 25.36 & 41.59 & 38.61 & 37.94 & 4.80 \\
 & Qwen3.5-9B & 26.20 & 26.16 & 26.24 & 26.00 & 40.10 & 40.31 & 38.15 & 5.80 \\
 & Qwen3.5-4B & 25.00 & 25.42 & 25.04 & 24.66 & 43.08 & 32.70 & 34.45 & 7.20 \\
\midrule
\multirow{5}{*}{T$^2$-Agent} & GPT-5.2 & 28.20 & 29.91 & 28.14 & 26.00 & 42.15 & 47.15 & 42.68 & 2.60 \\
 & Gemma4-31B & 24.40 & 24.97 & 24.41 & 24.60 & 41.16 & 38.73 & 38.73 & 7.60 \\
 & Qwen3.6-27B & 26.00 & 24.07 & 26.13 & 23.40 & 41.63 & 27.09 & 27.74 & 4.20 \\
 & Qwen3.5-9B & 25.60 & 23.98 & 25.67 & 23.31 & 38.83 & 27.14 & 28.58 & 3.20 \\
 & Qwen3.5-4B & 21.20 & 20.41 & 21.20 & 19.92 & 37.95 & 27.26 & 28.50 & 1.80 \\
\midrule
\multirow{5}{*}{ReMMD-Agent} & GPT-5.2 & \textbf{41.80} & \textbf{43.98} & \textbf{41.71} & \textbf{39.12} & 44.31 & 47.01 & 45.15 & 5.00 \\
 & Gemma4-31B & 33.60 & 34.21 & 33.59 & 33.76 & 43.67 & 41.31 & 42.27 & 7.80 \\
 & Qwen3.6-27B & 30.40 & 32.85 & 30.36 & 30.07 & \textbf{44.73} & 39.20 & 41.10 & \textbf{10.80} \\
 & Qwen3.5-9B & 37.20 & 39.72 & 37.11 & 37.18 & 44.58 & \textbf{50.88} & \textbf{46.97} & 10.00 \\
 & Qwen3.5-4B & 29.20 & 32.07 & 29.12 & 28.58 & 42.59 & 42.89 & 41.94 & 6.00 \\
\bottomrule
\end{tabular}
\caption{Full ReMMDBench results on 500 samples. Values are percentages. The grey rows mark general-purpose assistant baselines; each agent block merges five backbone rows.}
\label{tab:main-results}
\end{table*}

\subsection{Structured Evidence Judgment}

The final stage processes the parsed points alongside their corresponding retrieved evidence states. The structured judge first assigns a verification state to each atomic point $\sigma_j\in\{\mathrm{supported},\mathrm{contradicted},\mathrm{unverified}\}$, subsequently inferring the L1 veracity $y$ from the structural evidence pattern over central claims and cross-modal bindings. This step avoids naive voting mechanisms, recognizing that a contradicted peripheral number may merely shift a post to Mostly True, whereas a contradicted event attribution determines a False verdict regardless of authentic surface details. The L2 vector $z$ is assigned strictly after L1 to ensure visual provenance is evaluated as a diagnostic signal rather than a definitive shortcut for falsehood.

\subsection{Implementation Details}

Queries are primarily issued in the original language of the post. For cross-lingual samples, the verification pipeline actively expands its retrieval scope to target regional evidence native to the original geographical context of the source material, supplementing this process with an English or Chinese bridge query. Furthermore, the structured judge incorporates specific prompt directives to explicitly audit semantic discrepancies and translation-induced misalignments. Visual retrieval utilizes captions, OCR, named entities, and reverse-search descriptions when available. Auxiliary textual analysis flags fabrication, distortion, and misleading context, while visual analysis isolates synthetic content, editing traces, source mismatches, and cross-modal inconsistencies. Reproducibility resources, including the exact prompt templates, decoding configurations, failure-handling rules, and retrieval budgets, are described in Appendix~\ref{app:reproducibility-materials}.

\section{Experiments}

\subsection{Experimental Setup}

We evaluate Manus~\citep{manus2026}, ChatGPT~\citep{openai2026chatgpt}, MMD-Agent~\citep{liu2025mmfakebench}, T$^2$-Agent~\citep{cui2025t2agent}, and ReMMD-Agent on the full 500-sample ReMMDBench split. Manus uses Manus 1.6, and ChatGPT is evaluated through the OpenAI web interface. Model-backed agents use GPT-5.2, Gemma4-31B, Qwen3.6-27B, Qwen3.5-9B, and Qwen3.5-4B, with non-GPT open backbones deployed locally on H200 GPUs. All web retrieval uses the Serper API~\citep{serper2026}, and model-backed agents share the same evidence retriever and visual processing pipeline. We adapt MMD-Agent and T$^2$-Agent to accommodate multiple visual inputs while strictly preserving their original evaluation logic. Appendix~\ref{app:agent-adaptation} documents these structural adaptations, including image serialization, tool-budget allocation across concurrent visual claims, and standardized response parsing. Each system predicts the L1 five-way veracity label and L2 eight-label distortion vector. We report exact L1 accuracy and macro metrics, L2 macro metrics, and L2 exact match. GPT-5.2 cost is measured on the full benchmark under the same endpoint and tool-call budgets for all model-backed agents.

\subsection{Overall Results}

\begin{table}[ht]
\centering
\small
\setlength{\tabcolsep}{3.5pt}
\begin{tabular}{lrr}
\toprule
\textbf{Variant} & \textbf{L1 Macro-F1} & \textbf{L2 Macro-F1} \\
\midrule
full ReMMD-Agent & 39.12 & 45.15 \\
w/o memory bank & 35.84 & 41.77 \\
w/o atomic parsing & 34.96 & 39.88 \\
w/o visual auxiliary analysis & 37.21 & 40.46 \\
w/o search & 31.42 & 37.09 \\
\bottomrule
\end{tabular}
\caption{Ablation on the GPT-5.2 ReMMD-Agent. Atomic parsing and memory reuse both contribute, and visual auxiliary analysis is especially important for L2 labels.}
\label{tab:ablation}
\end{table}

Table~\ref{tab:main-results} shows that ReMMDBench remains difficult for all evaluated systems, which confirms that five-way, multi-image verification is substantially harder than detecting local suspicious cues. General-purpose assistants are competitive on some L2 metrics, but their weaker L1 results indicate that graded veracity depends on how evidence changes the central claim. ReMMD-Agent improves this calibration across backbone families. GPT-5.2 gives the best L1 performance, and Qwen3.5-9B gives the strongest L2 macro-F1 among comparable open-backbone runs.

The comparison with MMD-Agent and T$^2$-Agent shows that additional search is not sufficient unless evidence is organized around the right claims. MMD-Agent remains useful for distortion-oriented comparison, but struggles with partial-truth labels in long multi-image narratives. T$^2$-Agent explores more reasoning paths, yet the extra search does not consistently improve veracity. Appendix~\ref{app:benchmark-distributions} reports additional GPT-backed confusion matrices.

\begin{table}[t]
\centering
\small
\setlength{\tabcolsep}{4pt}
\begin{tabular}{lrrr}
\toprule
\textbf{System} & \textbf{Total} & \textbf{Per sample} & \textbf{vs. ReMMD} \\
\midrule
MMD-Agent & \$126.32 & \$0.2526 & 1.21$\times$ \\
T$^2$-Agent & \$517.91 & \$1.0358 & 4.97$\times$ \\
ReMMD-Agent & \$104.16 & \$0.2083 & 1.00$\times$ \\
\bottomrule
\end{tabular}
\caption{Full-benchmark GPT-5.2 cost audit. ReMMD-Agent reduces per-sample cost by 17.5\% relative to MMD-Agent and 79.9\% relative to T$^2$-Agent.}
\label{tab:cost}
\end{table}

Table~\ref{tab:ablation} identifies the mechanism behind these gains. Atomic parsing reduces long-form information noise and supplies checkable units for retrieval and diagnosis, while memory supports provenance aggregation and cross-image evidence reuse. Removing either component hurts both L1 and L2, and the single-pass judge is weakest. Visual auxiliary analysis is especially important for L2 because visual edits and cross-modal mismatches can be diagnostic before they determine the final veracity label.

\subsection{Fine-Grained Behavior}

\begin{figure*}[t]
  \centering
  \includegraphics[width=\textwidth]{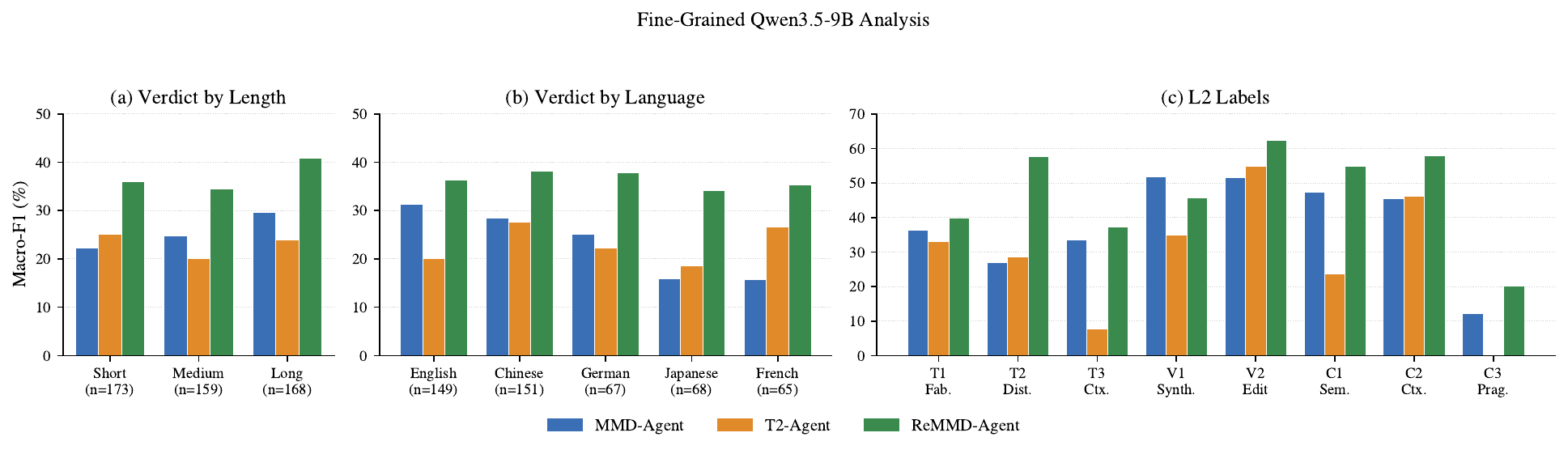}
  \caption{Fine-grained Qwen3.5-9B analysis across text length, language, and L2 labels.}
  \label{fig:fine-grained}
\end{figure*}

Figure~\ref{fig:fine-grained} tests whether the gains persist under the main pressures built into ReMMDBench. Across text-length tiers, ReMMD-Agent is more stable than the baselines. This is most informative for long posts, where additional context also introduces more entities, dates, quotations, and image references. Atomic parsing turns this noisy context into checkable units, and memory reuse reduces retrieval of superficially related but temporally or geographically mismatched events. Language slices show that multilingual verification is not only a translation problem, since regional source availability, entity grounding, and cross-script naming variation matter, especially for Japanese and French. Label slices show the clearest gains on distortion, editing, and cross-modal inconsistency labels, where evidence alignment is essential. The weaker advantage on synthetic visual content and pragmatic inconsistency suggests that low-level forensics and discourse-level support remain complementary challenges. Full numerical slices are reported in Appendix~\ref{app:fine-grained-results}.

\subsection{Cost and Transfer}

Table~\ref{tab:cost} demonstrates that performance gains emerge from architectural efficiency rather than increased computational expenditure. ReMMD-Agent achieves lower costs than MMD-Agent by systematically reusing evidence across atomic points and significantly undercuts T$^2$-Agent by eliminating the repeated expansion of tool-augmented reasoning paths. Such structural efficiency proves critical for dynamic benchmarks and high-concurrency real-world deployments requiring continuous verification. Furthermore, Table~\ref{tab:mmfake-transfer} explores the out-of-benchmark robustness of this evidence management paradigm. When deployed on the external MMFakeBench test split with a fixed Qwen3.5-9B backbone and uniform retrieval backend, ReMMD-Agent sustains substantial advantages over existing methods. While this binary fake detection setting confirms baseline algorithmic stability, it inherently simplifies the verification challenge, functioning as a preliminary indicator of structural robustness rather than fully establishing comprehensive transferability across complex multilingual environments, multi-image posts, five-way veracity spectra, or fine-grained distortion diagnoses. Appendix~\ref{app:mmfake-transfer} details this external evaluation configuration.

\begin{table}[t]
\centering
\small
\setlength{\tabcolsep}{4pt}
\begin{tabular}{lrr}
\toprule
\textbf{System} & \textbf{Accuracy} & \textbf{Macro-F1} \\
\midrule
MMD-Agent & 0.592 & 0.673 \\
T$^2$-Agent & 0.639 & 0.715 \\
ReMMD-Agent & \textbf{0.824} & \textbf{0.871} \\
\bottomrule
\end{tabular}
\caption{External out-of-benchmark evaluation on the MMFakeBench test split reporting binary accuracy and macro-F1 metrics under a unified Qwen3.5-9B backbone and retrieval configuration to provide a preliminary indicator of algorithmic robustness.}
\label{tab:mmfake-transfer}
\end{table}

\section{Discussion}

The main lesson is that realistic MMD is an evidence-selection problem. A post may use real evidence to support a wrong conclusion, so fine-grained labels are necessary. Retrieval helps only when each source is tied to the claim or image it verifies. Visual authenticity alone is not enough, because real images can be misused and synthetic images do not automatically falsify the text.

The Qwen results support this view. Under the same ReMMD-Agent pipeline, Qwen3.5-9B outperforms Qwen3.6-27B on several metrics. This is not a general reversal of model scale. After retrieval and memory provide evidence, the backbone mainly needs to follow the schema, calibrate partial evidence, and avoid over-interpreting uncertainty. Larger models can be less stable on adjacent partial-truth labels.

The benchmark also clarifies future directions. Rationales should identify the claim, evidence, and image-text relation. Multilingual cases require local entity and source grounding, not only translation. Future systems should improve source-aware memory, temporal retrieval, multilingual entity linking, and metrics that separately evaluate visual edits, verdicts, and misleading mechanisms.

\section{Conclusion}

We introduced ReMMDBench and ReMMD-Agent to study multimodal misinformation under realistic verification conditions. ReMMDBench moves evaluation beyond short binary image-text cases by combining multilingual posts, multiple images, graded veracity, distortion labels, and rationales. ReMMD-Agent shows that this setting is best handled as evidence management. It decomposes posts into checkable units, reuses retrieved evidence through memory, and judges veracity and distortion from an explicit evidence state. Experiments show that this design improves calibration, supports fine-grained distortion diagnosis, reduces retrieval cost, and transfers beyond ReMMDBench. Taken together, ReMMD reframes realistic multimodal misinformation detection around evidence selection, grounding, and explanation across modalities.

\bibliography{custom}

\clearpage
\appendix
\setcounter{secnumdepth}{1}

\begin{figure*}[tbp]
\centering
\includegraphics[width=0.95\textwidth]{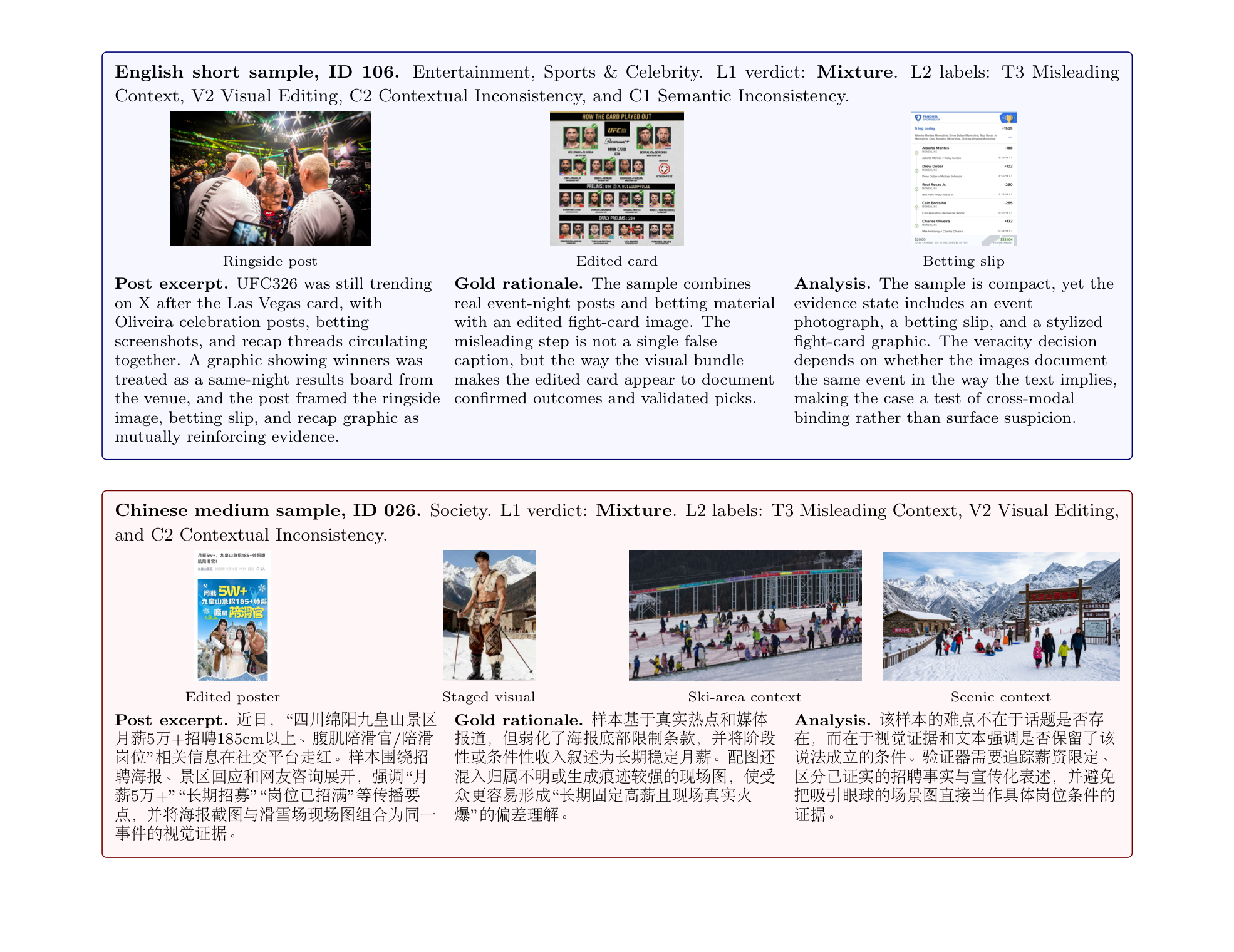}
\caption{Benchmark examples from ReMMDBench: an English short sample and a Chinese medium sample with multimodal evidence and distortion annotations.}
\label{fig:benchmark-examples}
\end{figure*}

\begin{figure*}[tbp]
\centering
\includegraphics[width=0.98\textwidth]{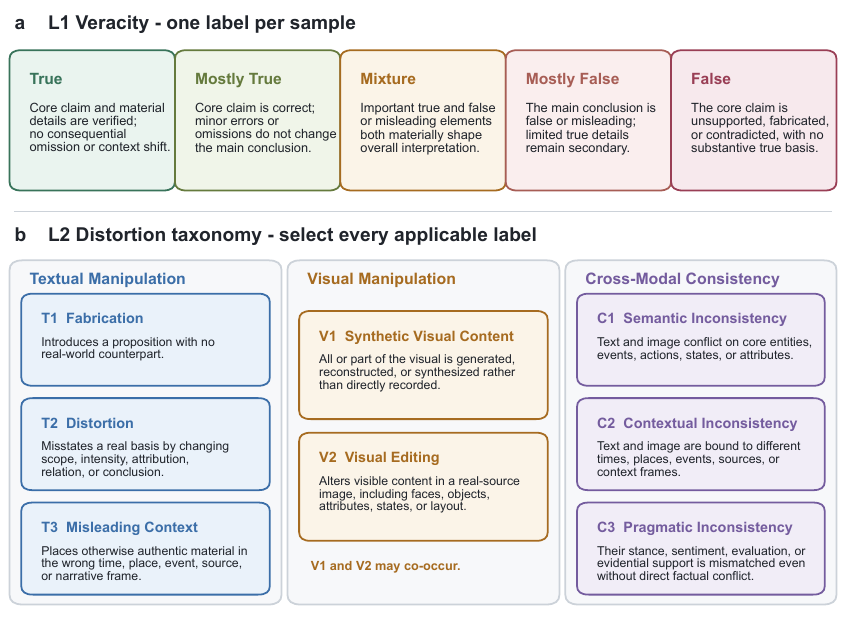}
\caption{Operational label definitions used in ReMMDBench. Panel (a) summarizes the five mutually exclusive L1 veracity verdicts according to the status of the core claim. Panel (b) summarizes the multi-label L2 taxonomy across textual manipulation, visual manipulation, and cross-modal consistency.}
\label{fig:appendix-label-definitions}
\end{figure*}

\section{Benchmark Examples}
\label{app:benchmark-examples}

Figure~\ref{fig:benchmark-examples} shows two non-sensitive ReMMDBench samples with multilingual text, multiple images, hierarchical labels, rationales, and evidence-centered analysis.

\section{Additional Benchmark Statistics}
\label{app:benchmark-statistics}

\paragraph{Data collection summary.}
The raw-data collection period ran from March 1, 2026, through May 10, 2026. Sources comprised Weibo, WeChat, X, the China Internet Joint Rumor Refuting Platform, FactCheck, and the Japan Fact-Check Center. The initial collection contained 54,468 posts; after screening and controlled rewriting, 500 benchmark samples were retained for ReMMDBench.

\paragraph{Analysis.}
Tables~\ref{tab:appendix-verdict-distribution} and~\ref{tab:appendix-distortion-frequency} separate the two annotation views that define ReMMDBench. The five L1 classes are close to balanced, which makes macro-F1 meaningful and prevents systems from succeeding by favoring a dominant verdict. The average number of L2 labels rises monotonically from True to False, showing that severe misinformation usually accumulates multiple forms of distortion rather than a single isolated cue. The distortion table further shows that visual editing, textual distortion, and cross-modal inconsistency all occur frequently. This distribution supports evaluating L1 veracity and L2 diagnosis together, since the same final verdict can arise from different combinations of textual, visual, and pragmatic evidence.

\section{Label Boundary Notes}
\label{app:label-boundaries}

T2 Distortion is assigned when a textual claim has a real factual basis but changes scope, intensity, attribution, relation, or conclusion. T3 Misleading Context is preferred when the content itself may be real but is placed in the wrong time, location, source, or event frame. V1 and V2 can co-occur when a real image is edited by inserting generated content. C1 concerns factual semantic conflict between text and image, C2 concerns context-frame mismatch, and C3 concerns stance, sentiment, or evidential-support mismatch.

Figure~\ref{fig:appendix-label-definitions} condenses the operational L1 and L2 definitions used during annotation. L1 assigns exactly one veracity verdict, whereas L2 records every applicable distortion type.

\section{Human--AI Annotation Agreement}
\label{app:annotation-agreement}

We assessed the GPT-5.4-based Annotator Agent across all 500 ReMMDBench samples, each of which was independently labeled by two professional human annotators. Figure~\ref{fig:appendix-label-definitions} defines the two output spaces evaluated here. For the ordered, mutually exclusive L1 labels, exact agreement measures identical verdicts, while linear-weighted Cohen's $\kappa$ additionally corrects for chance agreement and assigns smaller penalties to disagreements between adjacent severity levels. For the multi-label L2 taxonomy, exact set match requires the complete predicted and reference label sets to coincide, whereas mean Jaccard similarity credits partial set overlap.

\begin{table*}[tbp]
\centering
\caption{Agreement across all 500 ReMMDBench samples. Human 1 and Human 2 denote the two professional human annotators; Agent denotes the GPT-5.4-based Annotator Agent; and Human Consensus denotes their consensus labels. All metrics range from 0 to 1, and higher values indicate stronger agreement.}
\label{tab:annotation-agreement}
\footnotesize
\setlength{\tabcolsep}{3.5pt}
\renewcommand{\arraystretch}{1.20}
\begin{tabular}{lcccc}
\toprule
\textbf{Comparison} &
\shortstack{\textbf{L1 Exact}\\\textbf{Agreement} $\uparrow$} &
\shortstack{\textbf{L1 Linear-Weighted}\\$\boldsymbol{\kappa}$ $\uparrow$} &
\shortstack{\textbf{L2 Exact Set}\\\textbf{Match} $\uparrow$} &
\shortstack{\textbf{L2 Mean}\\\textbf{Jaccard} $\uparrow$} \\
\midrule
Human 1 vs. Human 2 & 0.96 & 0.97 & 0.83 & 0.94 \\
Agent vs. Human 1 & 0.93 & 0.93 & 0.79 & 0.91 \\
Agent vs. Human 2 & 0.89 & 0.91 & 0.76 & 0.85 \\
Agent vs. Human Consensus & 0.90 & 0.91 & 0.77 & 0.88 \\
\bottomrule
\end{tabular}
\end{table*}

\paragraph{Analysis.}
Table~\ref{tab:annotation-agreement} shows high reproducibility across the complete benchmark under the operational definitions in Figure~\ref{fig:appendix-label-definitions}. The two human annotators reached 0.96 exact agreement and 0.97 linear-weighted $\kappa$ on L1. Against the human consensus, the Annotator Agent retained 0.90 exact agreement and a weighted $\kappa$ of 0.91. The slightly higher weighted agreement is consistent with residual L1 disagreements being concentrated around nearby severity levels rather than arbitrary shifts across the five-way scale. L2 is stricter because a sample may receive several simultaneous distortion labels: human--human exact set match was 0.83, while Agent--consensus exact set match was 0.77. The corresponding mean Jaccard scores of 0.94 and 0.88 show that many non-exact cases still preserved substantial overlap in the selected distortion types. Together, these results establish strong agreement for the automated annotation pipeline while retaining a measurable gap from human--human reproducibility, particularly for complete L2 label sets.

\section{Agent Adaptation Details}
\label{app:agent-adaptation}

MMD-Agent and T$^2$-Agent were originally designed for single-image multimodal misinformation inputs. The following additional experimental configuration describes the implementation-specific input routing used to evaluate them on ReMMDBench. In both cases, the full post text is passed unchanged, and no ReMMDBench gold label or rationale is exposed during inference.

\paragraph{MMD-Agent multi-image input.}
Image files are resolved from the order listed in \texttt{sample.json}, deduplicated, and capped at 10 per sample. Each retained image is base64-encoded as a separate \texttt{image\_url} content block following the text prompt. The complete ordered image bundle is attached unchanged to every textual, visual, and cross-modal model call. The visual stage first produces a joint description of the image set; this description is inserted into the subsequent visual and cross-modal prompts, after which the original three-stage aggregation rule produces the L1 verdict and L2 labels.

\paragraph{T$^2$-Agent multi-image input.}
All image paths listed in \texttt{sample.json} are resolved in document order and retained. To preserve the original single-image MCTS and tool semantics, initialization, action planning, VQA, entity extraction, image detection, and counterfactual tool calls use the first image as their visual anchor. The text, image, and text--image matching subtrees therefore retain the original action space and trajectory structure. After tree search, their best trajectories and modality scores are serialized, and the complete ordered image set is attached to the final L2 taxonomy classifier; the L1 verdict remains determined by the three subtree probabilities.

\section{Reproducibility Materials and Repeated Runs}
\label{app:reproducibility-materials}

Complete code, exact prompt templates, decoding configurations, failure-handling rules, and retrieval budgets are included in the project repository. All experimental results reported in the paper are averages over three runs.

\section{MMFakeBench Transfer Setting}
\label{app:mmfake-transfer}

The transfer experiment uses the official 10,000-instance MMFakeBench test split, whose class distribution is 70\% fake and 30\% true. All compared agents use Qwen3.5-9B and the same retrieval backend (Google Search by Serper API). ReMMD-Agent obtains 0.824 accuracy and 0.871 Macro-F1. MMD-Agent obtains 0.592 accuracy and 0.673 Macro-F1, while T$^2$-Agent obtains 0.639 accuracy and 0.715 Macro-F1.

\paragraph{Analysis.}
The transfer setting reduces the output space from five-way veracity and eight distortion labels to binary fake detection. ReMMD-Agent still keeps a large advantage, which suggests that its benefit is not limited to ReMMDBench-specific label definitions. The result is also informative for smaller open-source backbones: T$^2$-Agent performs more search, but the additional reasoning loop does not compensate for weaker evidence routing when the backbone capacity is limited. ReMMD-Agent's decomposition and memory reuse appear to supply a more stable control policy under the same retriever.

\begin{table}[tbp]
\centering
\small
\begin{tabular}{lrr}
\toprule
\textbf{Verdict} & \textbf{Count} & \textbf{Avg. L2} \\
\midrule
True & 100 & 0.00 \\
Mostly True & 99 & 2.03 \\
Mixture & 100 & 3.52 \\
Mostly False & 102 & 3.93 \\
False & 99 & 4.41 \\
\bottomrule
\end{tabular}
\caption{Verdict distribution and average number of L2 labels.}
\label{tab:appendix-verdict-distribution}
\end{table}

\begin{table}[ht]
\centering
\small
\begin{tabular}{lrr}
\toprule
\textbf{Distortion label} & \textbf{Count} & \textbf{Percent} \\
\midrule
T1 Fabrication & 99 & 19.8 \\
T2 Distortion & 222 & 44.4 \\
T3 Misleading Context & 164 & 32.8 \\
V1 Synthetic Visual Content & 145 & 29.0 \\
V2 Visual Editing & 272 & 54.4 \\
C1 Semantic Inconsistency & 212 & 42.4 \\
C2 Contextual Inconsistency & 210 & 42.0 \\
C3 Pragmatic Inconsistency & 67 & 13.4 \\
\bottomrule
\end{tabular}
\caption{Distortion-label frequency in ReMMDBench.}
\label{tab:appendix-distortion-frequency}
\end{table}

\section{Additional Fine-Grained Results}
\label{app:fine-grained-results}

The following tables report the complete fine-grained slices used in the main paper's fine-grained analysis. Tables~\ref{tab:appendix-short-results}, \ref{tab:appendix-medium-results}, and~\ref{tab:appendix-long-results} use the same grouped-agent layout as the main results. Grey rows denote general-purpose assistant baselines, and agent rows are grouped by system family.

\begin{table*}[tbp]
\centering
\small
\setlength{\tabcolsep}{2.2pt}
\begin{tabular}{llrrrrrrrr}
\toprule
\multirow{2}{*}{\textbf{Agent}} & \multirow{2}{*}{\textbf{Backbone}} & \multicolumn{4}{c}{\textbf{L1 Veracity}} & \multicolumn{4}{c}{\textbf{L2 Distortion}} \\
\cmidrule(lr){3-6}\cmidrule(lr){7-10}
 & & \textbf{Accuracy} & \textbf{Precision} & \textbf{Recall} & \textbf{Macro-F1} & \textbf{Precision} & \textbf{Recall} & \textbf{Macro-F1} & \textbf{Exact Match} \\
\midrule
Manus & proprietary & 30.06 & 30.31 & 29.89 & 29.84 & 40.06 & 41.08 & 40.23 & 6.36 \\
ChatGPT & proprietary & 31.21 & 25.23 & 30.67 & 27.21 & 37.73 & 45.99 & 40.74 & 1.73 \\
\midrule
\multirow{5}{*}{MMD-Agent} & GPT-5.2 & 26.59 & 26.28 & 26.63 & 24.10 & 43.41 & 42.25 & 38.86 & 2.89 \\
 & Gemma4-31B & 24.86 & 26.31 & 24.98 & 24.67 & 38.59 & 42.21 & 39.66 & 5.20 \\
 & Qwen3.6-27B & 23.12 & 24.22 & 23.43 & 23.10 & 41.83 & 35.07 & 36.20 & 6.94 \\
 & Qwen3.5-9B & 23.12 & 22.16 & 22.97 & 22.18 & 40.02 & 40.27 & 37.77 & 5.20 \\
 & Qwen3.5-4B & 28.90 & 30.89 & 28.80 & 28.17 & 40.09 & 34.00 & 34.05 & 8.67 \\
\midrule
\multirow{5}{*}{T$^2$-Agent} & GPT-5.2 & 26.59 & 34.55 & 26.40 & 24.73 & 40.51 & 45.47 & 40.74 & 2.31 \\
 & Gemma4-31B & 25.43 & 26.24 & 25.54 & 25.83 & 41.55 & 40.98 & 40.15 & 9.83 \\
 & Qwen3.6-27B & 30.06 & 27.91 & 29.29 & 26.13 & 41.63 & 27.70 & 29.75 & 5.78 \\
 & Qwen3.5-9B & 28.90 & 25.08 & 27.96 & 25.09 & 37.63 & 26.36 & 27.59 & 2.89 \\
 & Qwen3.5-4B & 26.01 & 23.35 & 25.05 & 23.16 & 37.06 & 28.29 & 28.49 & 2.31 \\
\midrule
\multirow{5}{*}{ReMMD-Agent} & GPT-5.2 & \textbf{41.04} & \textbf{40.11} & \textbf{40.51} & \textbf{37.91} & 41.74 & 48.36 & \textbf{44.45} & 5.78 \\
 & Gemma4-31B & 35.26 & 35.17 & 35.07 & 34.85 & 41.10 & 41.32 & 40.91 & 6.94 \\
 & Qwen3.6-27B & 31.79 & 33.76 & 31.88 & 31.14 & 43.05 & 36.77 & 39.01 & \textbf{9.83} \\
 & Qwen3.5-9B & 36.42 & 38.96 & 36.29 & 36.07 & 40.86 & 43.43 & 41.49 & \textbf{9.83} \\
 & Qwen3.5-4B & 32.95 & 35.18 & 32.81 & 32.62 & 39.95 & 37.66 & 37.95 & 6.36 \\
\bottomrule
\end{tabular}
\caption{Short-text subset results on 173 samples. Values are percentages and the table follows the same layout as the main results.}
\label{tab:appendix-short-results}
\end{table*}

\paragraph{Short-text analysis.}
Short posts contain fewer claims and fewer images, but they provide less context for disambiguating entities and events. ReMMD-Agent still leads the strongest L1 results, especially with GPT-5.2, indicating that atomic decomposition is useful even when the textual input is compact. The L2 gap is smaller than in longer tiers because many distortions are visible from local cues, which lets assistant-style baselines remain competitive. Even so, exact match remains low across systems. This indicates that short posts often compress several cues into a small space, so a system must still decide whether a visual cue changes the central claim or only adds suspicious context.

\begin{table*}[tbp]
\centering
\small
\setlength{\tabcolsep}{2.2pt}
\begin{tabular}{llrrrrrrrr}
\toprule
\multirow{2}{*}{\textbf{Agent}} & \multirow{2}{*}{\textbf{Backbone}} & \multicolumn{4}{c}{\textbf{L1 Veracity}} & \multicolumn{4}{c}{\textbf{L2 Distortion}} \\
\cmidrule(lr){3-6}\cmidrule(lr){7-10}
 & & \textbf{Accuracy} & \textbf{Precision} & \textbf{Recall} & \textbf{Macro-F1} & \textbf{Precision} & \textbf{Recall} & \textbf{Macro-F1} & \textbf{Exact Match} \\
\midrule
Manus & proprietary & 29.56 & 30.38 & 28.98 & 29.34 & 44.57 & 46.41 & 44.83 & 8.18 \\
ChatGPT & proprietary & 25.16 & 32.05 & 25.97 & 24.48 & 40.21 & 46.86 & 41.97 & 4.40 \\
\midrule
\multirow{5}{*}{MMD-Agent} & GPT-5.2 & 23.27 & 18.02 & 23.21 & 19.13 & 38.56 & 49.06 & 39.43 & 2.52 \\
 & Gemma4-31B & 23.90 & 24.53 & 23.26 & 23.61 & 40.72 & 41.24 & 40.08 & 5.66 \\
 & Qwen3.6-27B & 30.82 & 32.68 & 30.56 & 30.98 & 39.41 & 38.43 & 37.59 & 6.29 \\
 & Qwen3.5-9B & 26.42 & 25.28 & 25.23 & 24.84 & 38.49 & 36.58 & 35.93 & 9.43 \\
 & Qwen3.5-4B & 25.79 & 24.49 & 25.30 & 24.69 & 41.24 & 31.35 & 33.41 & 6.92 \\
\midrule
\multirow{5}{*}{T$^2$-Agent} & GPT-5.2 & 28.30 & 23.57 & 29.86 & 25.92 & 37.94 & 46.03 & 39.82 & 1.89 \\
 & Gemma4-31B & 25.16 & 26.33 & 24.97 & 25.08 & 35.52 & 35.35 & 33.96 & 8.18 \\
 & Qwen3.6-27B & 25.79 & 24.01 & 25.31 & 23.54 & 40.00 & 26.93 & 28.10 & 3.14 \\
 & Qwen3.5-9B & 22.64 & 20.49 & 22.70 & 20.03 & 34.08 & 26.81 & 27.65 & 3.77 \\
 & Qwen3.5-4B & 19.50 & 19.91 & 20.12 & 19.04 & 36.60 & 25.66 & 27.71 & 1.26 \\
\midrule
\multirow{5}{*}{ReMMD-Agent} & GPT-5.2 & \textbf{40.25} & \textbf{48.31} & \textbf{41.31} & \textbf{39.01} & 45.87 & 49.55 & 47.00 & 5.66 \\
 & Gemma4-31B & 33.33 & 33.09 & 32.76 & 32.10 & 41.44 & 40.50 & 40.61 & 10.69 \\
 & Qwen3.6-27B & 30.19 & 33.28 & 29.62 & 29.59 & \textbf{46.03} & 38.70 & 41.29 & \textbf{12.58} \\
 & Qwen3.5-9B & 33.96 & 36.08 & 34.79 & 34.55 & 45.13 & \textbf{52.62} & \textbf{47.70} & 11.32 \\
 & Qwen3.5-4B & 22.64 & 26.79 & 22.95 & 22.50 & 40.56 & 43.98 & 41.39 & 6.92 \\
\bottomrule
\end{tabular}
\caption{Medium-text subset results on 159 samples. Values are percentages and the table follows the same layout as the main results.}
\label{tab:appendix-medium-results}
\end{table*}

\paragraph{Medium-text analysis.}
The medium tier is where simple scaling of context begins to fail. Several baselines improve L2 recall, but their L1 macro-F1 remains unstable because partial evidence must be assigned to the correct severity class. ReMMD-Agent/GPT-5.2 gives the strongest verdict performance, while ReMMD-Agent/Qwen3.5-9B gives the best L2 macro-F1. This split suggests that larger proprietary backbones help with calibrated verdict assignment, whereas the decomposition policy can still help a smaller open-source model detect distortion mechanisms. The tier is therefore diagnostic of the benchmark's main difficulty: additional narrative context creates more opportunities for evidence retrieval, but also increases the risk of treating peripheral contradictions as central.

\paragraph{Long-text analysis.}
The long tier is the most realistic stress test because the average sample contains about ten images and a much longer narrative. ReMMD-Agent/Qwen3.5-9B reaches the highest L1 macro-F1 and L2 macro-F1 in this slice, while ReMMD-Agent/GPT-5.2 has the highest accuracy. This pattern indicates that long posts reward evidence organization: additional context helps only when the agent can bind claims, images, and retrieved sources. The weaker T$^2$-Agent results show that expanding the reasoning search space is not enough if the retrieved evidence is not tied back to stable atomic units. Long posts also magnify the difference between retrieval volume and retrieval usefulness, since many plausible sources may describe neighboring events, reused images, or partially matching entities.

\begin{table*}[tbp]
\centering
\small
\setlength{\tabcolsep}{2.2pt}
\begin{tabular}{llrrrrrrrr}
\toprule
\multirow{2}{*}{\textbf{Agent}} & \multirow{2}{*}{\textbf{Backbone}} & \multicolumn{4}{c}{\textbf{L1 Veracity}} & \multicolumn{4}{c}{\textbf{L2 Distortion}} \\
\cmidrule(lr){3-6}\cmidrule(lr){7-10}
 & & \textbf{Accuracy} & \textbf{Precision} & \textbf{Recall} & \textbf{Macro-F1} & \textbf{Precision} & \textbf{Recall} & \textbf{Macro-F1} & \textbf{Exact Match} \\
\midrule
Manus & proprietary & 39.29 & 39.02 & 38.91 & 38.84 & 46.92 & 40.52 & 43.23 & 8.33 \\
ChatGPT & proprietary & 33.93 & 34.35 & 32.99 & 31.90 & 47.74 & 49.76 & 47.83 & 2.98 \\
\midrule
\multirow{5}{*}{MMD-Agent} & GPT-5.2 & 29.17 & 26.42 & 28.53 & 25.82 & 44.28 & 54.67 & 46.18 & 0.60 \\
 & Gemma4-31B & 27.98 & 28.41 & 27.92 & 27.65 & 43.30 & 42.95 & 42.01 & 4.76 \\
 & Qwen3.6-27B & 22.62 & 23.11 & 22.73 & 22.29 & 44.04 & 41.75 & 38.99 & 1.19 \\
 & Qwen3.5-9B & 29.17 & 30.36 & 29.33 & 29.66 & 41.79 & 43.80 & 39.82 & 2.98 \\
 & Qwen3.5-4B & 20.24 & 21.46 & 20.91 & 20.43 & 48.59 & 32.89 & 35.70 & 5.95 \\
\midrule
\multirow{5}{*}{T$^2$-Agent} & GPT-5.2 & 29.76 & 31.51 & 29.14 & 27.65 & 47.89 & 49.98 & 46.90 & 3.57 \\
 & Gemma4-31B & 22.62 & 22.39 & 22.40 & 22.31 & 45.90 & 39.20 & 40.95 & 4.76 \\
 & Qwen3.6-27B & 22.02 & 19.35 & 23.74 & 19.20 & 38.81 & 26.54 & 25.18 & 3.57 \\
 & Qwen3.5-9B & 25.00 & 24.96 & 25.90 & 23.95 & 43.80 & 28.18 & 29.88 & 2.98 \\
 & Qwen3.5-4B & 17.86 & 17.09 & 18.34 & 16.98 & 40.67 & 27.51 & 28.89 & 1.79 \\
\midrule
\multirow{5}{*}{ReMMD-Agent} & GPT-5.2 & \textbf{44.05} & 41.18 & \textbf{42.78} & 39.41 & 45.36 & 42.71 & 43.37 & 3.57 \\
 & Gemma4-31B & 32.14 & 32.19 & 32.33 & 31.75 & 49.36 & 41.78 & 44.89 & 5.95 \\
 & Qwen3.6-27B & 29.17 & 30.86 & 29.09 & 28.46 & 45.00 & 41.85 & 42.59 & \textbf{10.12} \\
 & Qwen3.5-9B & 41.07 & \textbf{46.94} & 40.24 & \textbf{40.79} & 46.88 & \textbf{55.97} & \textbf{50.43} & 8.93 \\
 & Qwen3.5-4B & 31.55 & 34.43 & 29.82 & 29.00 & 46.42 & 46.44 & 45.51 & 4.76 \\
\bottomrule
\end{tabular}
\caption{Long-text subset results on 168 samples. Values are percentages and the table follows the same layout as the main results.}
\label{tab:appendix-long-results}
\end{table*}

\begin{table*}[tbp]
\centering
\small
\setlength{\tabcolsep}{4pt}
\begin{tabular}{llrrrr}
\toprule
\textbf{Language} & \textbf{System} & \textbf{L1 F1} & \textbf{L2 F1} & \textbf{$\Delta$L1 vs. MMD} & \textbf{$\Delta$L2 vs. MMD} \\
\midrule
English & MMD-Agent/Qwen3.5-9B & 31.22 & 34.27 & ref. & ref. \\
English & T$^2$-Agent/Qwen3.5-9B & 20.09 & 27.45 & -11.13 & -6.82 \\
English & ReMMD-Agent/Qwen3.5-9B & \textbf{36.30} & \textbf{47.76} & \textbf{+5.08} & \textbf{+13.49} \\
\midrule
Chinese & MMD-Agent/Qwen3.5-9B & 28.51 & 44.30 & ref. & ref. \\
Chinese & T$^2$-Agent/Qwen3.5-9B & 27.62 & 33.22 & -0.90 & -11.07 \\
Chinese & ReMMD-Agent/Qwen3.5-9B & \textbf{38.20} & \textbf{48.81} & \textbf{+9.68} & \textbf{+4.51} \\
\midrule
German & MMD-Agent/Qwen3.5-9B & 25.11 & 39.23 & ref. & ref. \\
German & T$^2$-Agent/Qwen3.5-9B & 22.29 & 30.35 & -2.82 & -8.88 \\
German & ReMMD-Agent/Qwen3.5-9B & \textbf{37.78} & \textbf{48.41} & \textbf{+12.67} & \textbf{+9.18} \\
\midrule
Japanese & MMD-Agent/Qwen3.5-9B & 15.97 & 30.58 & ref. & ref. \\
Japanese & T$^2$-Agent/Qwen3.5-9B & 18.56 & 21.02 & +2.60 & -9.57 \\
Japanese & ReMMD-Agent/Qwen3.5-9B & \textbf{34.14} & \textbf{44.41} & \textbf{+18.18} & \textbf{+13.82} \\
\midrule
French & MMD-Agent/Qwen3.5-9B & 15.66 & 33.47 & ref. & ref. \\
French & T$^2$-Agent/Qwen3.5-9B & 26.68 & 25.27 & +11.02 & -8.20 \\
French & ReMMD-Agent/Qwen3.5-9B & \textbf{35.28} & \textbf{38.28} & \textbf{+19.62} & \textbf{+4.81} \\
\bottomrule
\end{tabular}%
\caption{Language-slice results for the Qwen3.5-9B backbone. The table reports verdict macro-F1 and distortion macro-F1, together with absolute gains over MMD-Agent.}
\label{tab:appendix-language-results}
\end{table*}

\paragraph{Language-slice analysis.}
Table~\ref{tab:appendix-language-results} shows that ReMMD-Agent improves both verdict and distortion performance in every language. The gains are largest for Japanese and French on L1, where MMD-Agent is weakest, suggesting that multilingual verification is constrained by entity anchoring and regional evidence access rather than translation alone. T$^2$-Agent occasionally improves L1 over MMD-Agent, as in French, but its L2 performance drops sharply. This indicates that broader search may find enough evidence for a coarse verdict while still failing to diagnose the distortion mechanism. The consistent L2 gains are especially important because distortion labels require matching local expressions, named entities, and media provenance across languages, not merely translating the post into English.

\begin{table*}[tbp]
\centering
\small
\setlength{\tabcolsep}{4pt}
\begin{tabular}{llrrrr}
\toprule
\textbf{Label} & \textbf{System} & \textbf{F1} & \textbf{$\Delta$ vs. MMD} & \textbf{$\Delta$ vs. T$^2$} & \textbf{Best} \\
\midrule
T1 Fabrication & MMD-Agent/Qwen3.5-9B & 36.43 & ref. & +3.43 &  \\
T1 Fabrication & T$^2$-Agent/Qwen3.5-9B & 33.00 & -3.43 & ref. &  \\
T1 Fabrication & ReMMD-Agent/Qwen3.5-9B & \textbf{39.80} & \textbf{+3.37} & \textbf{+6.80} & \cmark \\
\midrule
T2 Distortion & MMD-Agent/Qwen3.5-9B & 27.04 & ref. & -1.43 &  \\
T2 Distortion & T$^2$-Agent/Qwen3.5-9B & 28.47 & +1.43 & ref. &  \\
T2 Distortion & ReMMD-Agent/Qwen3.5-9B & \textbf{57.75} & \textbf{+30.71} & \textbf{+29.27} & \cmark \\
\midrule
T3 Misleading Context & MMD-Agent/Qwen3.5-9B & 33.62 & ref. & +26.01 &  \\
T3 Misleading Context & T$^2$-Agent/Qwen3.5-9B & 7.61 & -26.01 & ref. &  \\
T3 Misleading Context & ReMMD-Agent/Qwen3.5-9B & \textbf{37.29} & \textbf{+3.66} & \textbf{+29.68} & \cmark \\
\midrule
V1 Synthetic Visual Content & MMD-Agent/Qwen3.5-9B & \textbf{51.78} & ref. & \textbf{+16.80} & \cmark \\
V1 Synthetic Visual Content & T$^2$-Agent/Qwen3.5-9B & 34.98 & -16.80 & ref. &  \\
V1 Synthetic Visual Content & ReMMD-Agent/Qwen3.5-9B & 45.63 & -6.14 & +10.65 &  \\
\midrule
V2 Visual Editing & MMD-Agent/Qwen3.5-9B & 51.60 & ref. & -3.13 &  \\
V2 Visual Editing & T$^2$-Agent/Qwen3.5-9B & 54.73 & +3.13 & ref. &  \\
V2 Visual Editing & ReMMD-Agent/Qwen3.5-9B & \textbf{62.29} & \textbf{+10.69} & \textbf{+7.56} & \cmark \\
\midrule
C1 Semantic Inconsistency & MMD-Agent/Qwen3.5-9B & 47.24 & ref. & +23.66 &  \\
C1 Semantic Inconsistency & T$^2$-Agent/Qwen3.5-9B & 23.57 & -23.66 & ref. &  \\
C1 Semantic Inconsistency & ReMMD-Agent/Qwen3.5-9B & \textbf{54.88} & \textbf{+7.65} & \textbf{+31.31} & \cmark \\
\midrule
C2 Contextual Inconsistency & MMD-Agent/Qwen3.5-9B & 45.41 & ref. & -0.85 &  \\
C2 Contextual Inconsistency & T$^2$-Agent/Qwen3.5-9B & 46.26 & +0.85 & ref. &  \\
C2 Contextual Inconsistency & ReMMD-Agent/Qwen3.5-9B & \textbf{57.86} & \textbf{+12.45} & \textbf{+11.60} & \cmark \\
\midrule
C3 Pragmatic Inconsistency & MMD-Agent/Qwen3.5-9B & 12.07 & ref. & +12.07 &  \\
C3 Pragmatic Inconsistency & T$^2$-Agent/Qwen3.5-9B & 0.00 & -12.07 & ref. &  \\
C3 Pragmatic Inconsistency & ReMMD-Agent/Qwen3.5-9B & \textbf{20.21} & \textbf{+8.14} & \textbf{+20.21} & \cmark \\
\bottomrule
\end{tabular}%
\caption{Per-label L2 F1 for the Qwen3.5-9B backbone. The only label where ReMMD-Agent is not best is V1, indicating that low-level synthetic-image cues remain complementary to evidence retrieval.}
\label{tab:appendix-label-results}
\end{table*}

\paragraph{Distortion-label analysis.}
Table~\ref{tab:appendix-label-results} confirms that ReMMD-Agent is strongest on labels that require evidence alignment, especially T2 Distortion, V2 Visual Editing, C1 Semantic Inconsistency, and C2 Contextual Inconsistency. These labels depend on comparing the post with external evidence or with the intended image-text binding. The exception is V1 Synthetic Visual Content, where MMD-Agent performs best, suggesting that low-level generation artifacts and forensic cues remain useful even when retrieval is strong. C3 Pragmatic Inconsistency remains difficult for all systems because it depends on the rhetorical use of evidence rather than a single factual contradiction. This pattern supports the paper's central design choice: retrieval memory and atomic parsing are most valuable when the task is to decide how an otherwise plausible source is being used.

\section{Additional Benchmark Distributions and Confusion Matrices}
\label{app:benchmark-distributions}

\begin{figure*}[tbp]
  \centering
  \includegraphics[width=\textwidth]{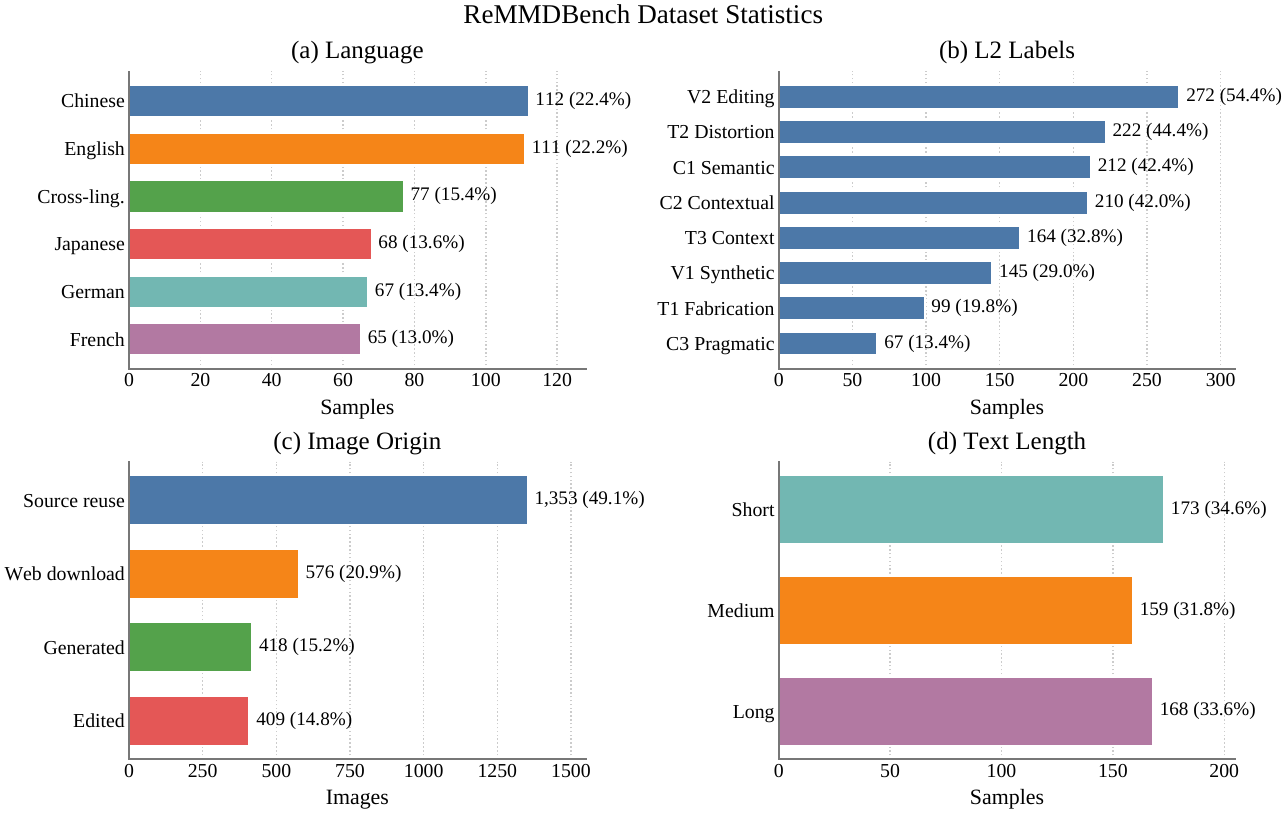}
  \caption{ReMMDBench statistics over language, L2 distortion labels, image provenance, and text-length tiers. Each panel reports counts with percentages in the corresponding sample or image population.}
  \label{fig:bench-stats-panels}
\end{figure*}

\paragraph{Benchmark-distribution analysis.}
Figure~\ref{fig:bench-stats-panels} summarizes the design pressures behind ReMMDBench. The language panel shows that the benchmark is not English-centric and includes cross-lingual cases as a distinct condition. The distortion panel confirms that textual, visual, and cross-modal labels all occur frequently, so systems cannot optimize for a single manipulation family. The provenance panel shows that the dataset mixes reused source images, web-downloaded evidence images, generated images, and edited images. The length panel verifies that short, medium, and long posts are balanced, which supports the main paper's length-tier analysis. Together, these distributions make the benchmark resistant to narrow shortcuts: a system must handle language variation, visual provenance, and text-image binding at the same time.

\begin{figure*}[tbp]
  \centering
  \includegraphics[width=0.9\linewidth]{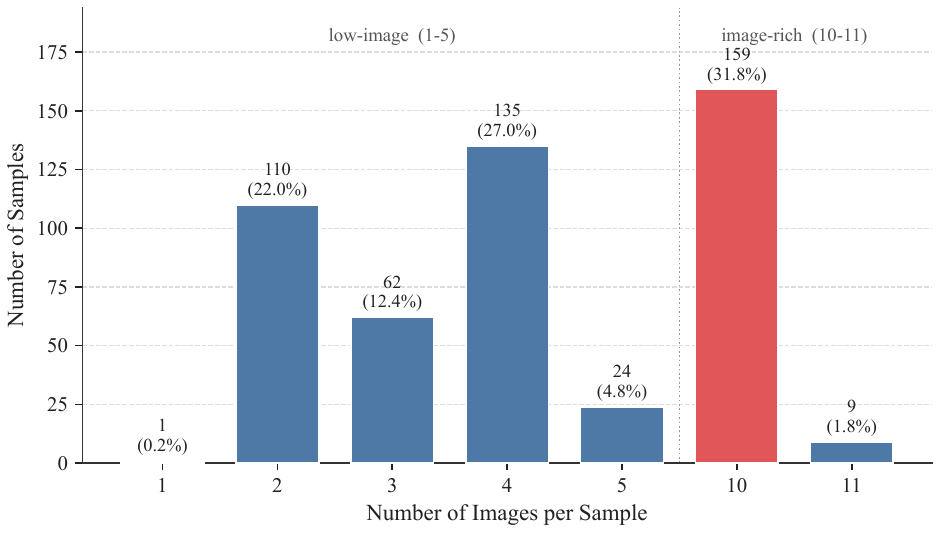}
  \caption{Distribution of images per sample in ReMMDBench. The long tail toward ten or eleven images is intentional and tests whether agents can aggregate evidence across carousel-style posts.}
  \label{fig:appendix-images-per-sample}
\end{figure*}

\paragraph{Image-count analysis.}
Figure~\ref{fig:appendix-images-per-sample} shows a long tail toward ten and eleven images. This shape is intentional rather than incidental. Many real social-media posts use carousel-style evidence, where some images are central and others are decorative, repeated, or weakly related. A verifier must therefore identify which images actually support the claim and which only add persuasive context. This is one reason ReMMDBench is difficult for agents that treat the image set as an undifferentiated visual bundle. The distribution also explains why memory reuse matters: once evidence is retrieved for one image or claim, it can often resolve later bindings without repeating the same search.

\begin{figure*}[tbp]
  \centering
  \includegraphics[width=0.98\textwidth]{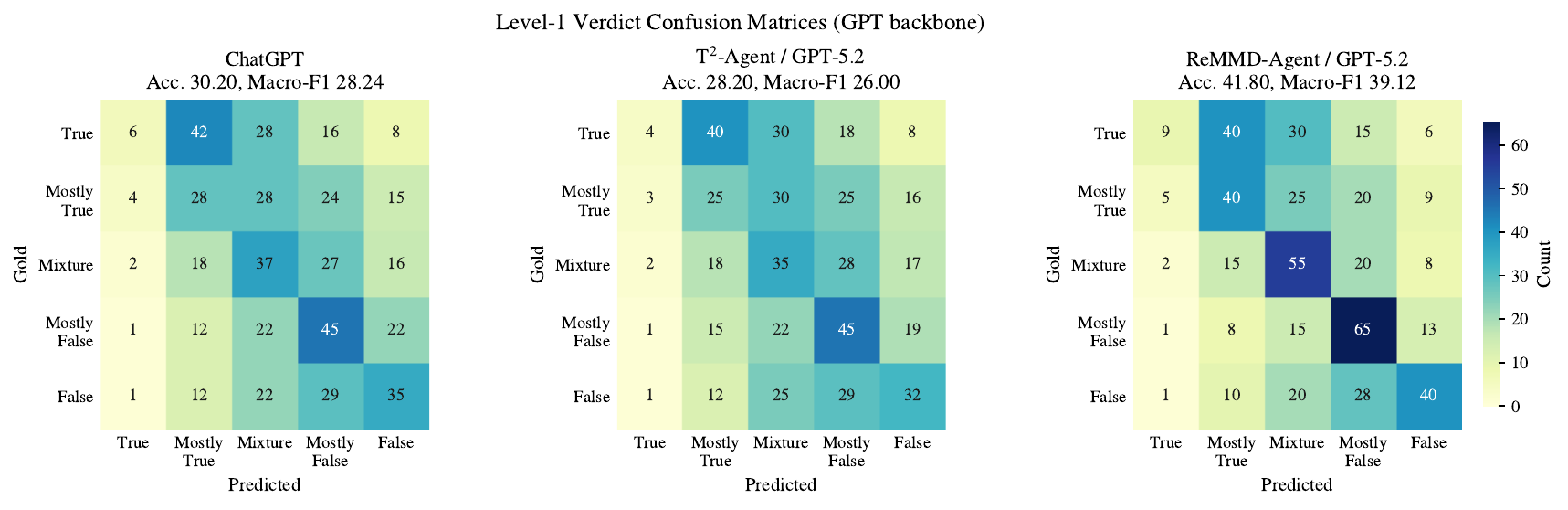}
  \caption{Appendix L1 count heatmaps for GPT-backed systems. Direct prompting and T$^2$-Agent often drift toward neighboring middle classes, while ReMMD-Agent recovers more diagonal mass without eliminating the intrinsic ambiguity of partial-truth cases.}
  \label{fig:appendix-gpt-confusion}
\end{figure*}

\paragraph{Confusion-matrix analysis.}
Figure~\ref{fig:appendix-gpt-confusion} compares GPT-backed systems under the five-way verdict scale. Direct prompting and T$^2$-Agent show a visible tendency to avoid confident True predictions and to concentrate mass around middle labels. This suggests a conservative model bias: when the task involves misinformation, models often treat uncertainty itself as evidence of partial falsehood. ReMMD-Agent reduces this drift by forcing the judge to keep supported, contradicted, and unverified atomic points separate. The remaining confusion around Mostly True, Mixture, and Mostly False is expected, because these labels depend on the centrality of the disputed evidence rather than the mere presence of an error. The matrix therefore provides a qualitative explanation for the macro-F1 gains: the agent improves not by eliminating ambiguity, but by reducing systematic drift caused by unmanaged uncertainty.

\clearpage
\begin{figure*}[t]
  \centering
  \includegraphics[width=\textwidth]{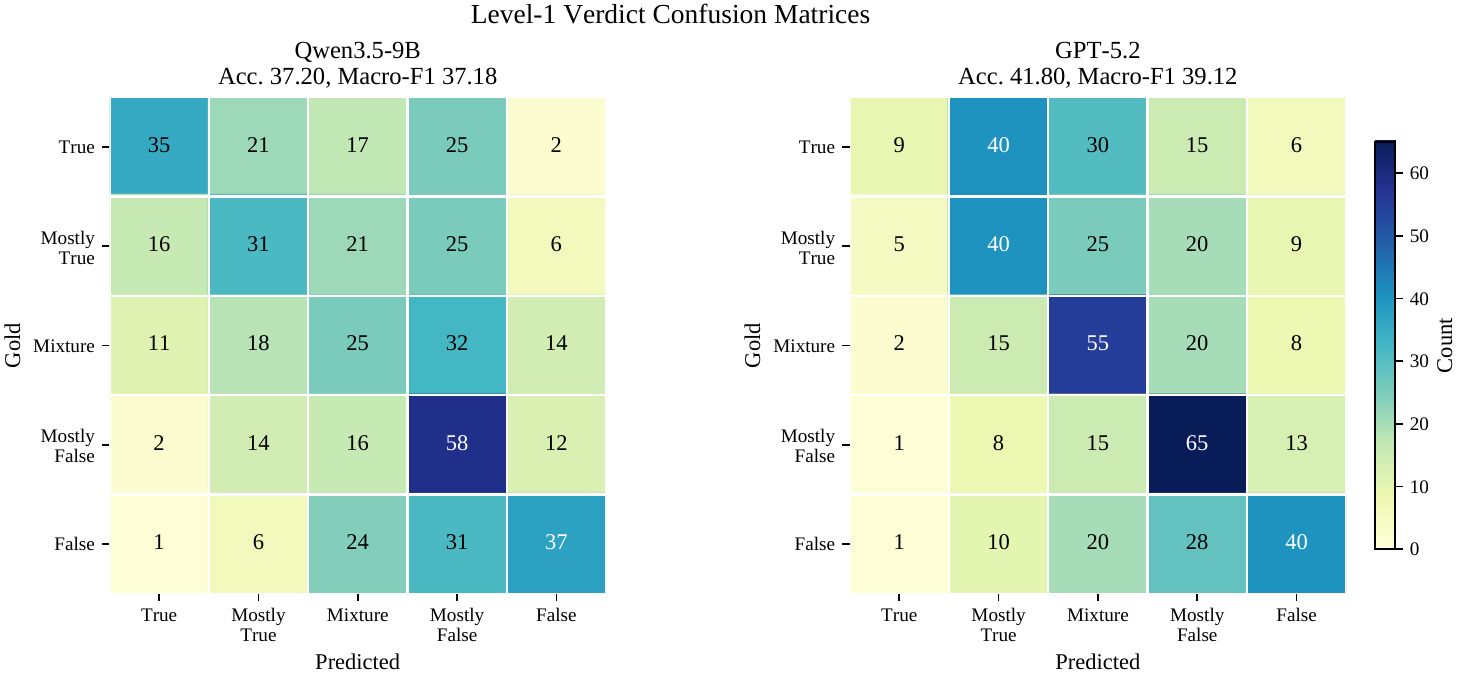}
  \caption{Count heatmaps for ReMMD-Agent L1 predictions. Both backbones recover substantial diagonal mass, but errors concentrate around adjacent middle labels where partial evidence must be calibrated rather than merely detected.}
  \label{fig:main-confusion}
  \caption*{Figure~\ref{fig:main-confusion} further shows that the remaining errors concentrate among neighboring middle labels, where models must judge the centrality of contradicted evidence rather than merely detect a suspicious cue.}
\end{figure*}

\end{document}